\documentclass{ieeeaccess}
\usepackage{cite}
\usepackage{amsmath,amssymb,amsfonts}
\usepackage{algorithmic}
\usepackage{graphicx}
\usepackage{textcomp}
\usepackage{tabularx}
\usepackage{booktabs}
\usepackage{hyperref}

\usepackage{bm}
\makeatletter
\AtBeginDocument{\DeclareMathVersion{bold}
\SetSymbolFont{operators}{bold}{T1}{times}{b}{n}
\SetSymbolFont{NewLetters}{bold}{T1}{times}{b}{it}
\SetMathAlphabet{\mathrm}{bold}{T1}{times}{b}{n}
\SetMathAlphabet{\mathit}{bold}{T1}{times}{b}{it}
\SetMathAlphabet{\mathbf}{bold}{T1}{times}{b}{n}
\SetMathAlphabet{\mathtt}{bold}{OT1}{pcr}{b}{n}
\SetSymbolFont{symbols}{bold}{OMS}{cmsy}{b}{n}
\renewcommand\boldmath{\@nomath\boldmath\mathversion{bold}}}
\makeatother

\def\BibTeX{{\rm B\kern-.05em{\sc i\kern-.025em b}\kern-.08em
    T\kern-.1667em\lower.7ex\hbox{E}\kern-.125emX}}

\begin{document}
\history{Date of submission Aug 16, 2024, date of current version Aug 16, 2024.}
\doi{10.1109/ACCESS.2024.0031879}

\def\thevol{8}
\def\theyear{2024}

\title{Enhancing Ship Classification in Optical Satellite Imagery: Integrating Convolutional Block Attention Module with ResNet for Improved Performance}

\author{
\uppercase{Ryan Donghan Kwon}\authorrefmark{1}, \IEEEmembership{Member, IEEE},
\uppercase{Gangjoo Robin Nam}\authorrefmark{2},\\
\uppercase{Jisoo Tak}\authorrefmark{1},
\uppercase{Junseob Shin}\authorrefmark{3},
\uppercase{Hyerin Cha}\authorrefmark{4} and
\uppercase{Seung Won Lee}\authorrefmark{1}
}

\address[1]{Sungkyunkwan University School of Medicine, Gyeonggi-do, 16419 South Korea}
\address[2]{Department of Research and Development, The Coala Inc., Seoul, 06190 South Korea}
\address[3]{Korea Science Academy of Korea Advanced Institute of Science and Technology, Busan, 47162 South Korea}
\address[4]{Seoul National University School of Earth and Environmental Sciences, Seoul, 08826 South Korea}

\corresp{Corresponding author: Ryan Donghan Kwon and Seung Won Lee (e-mail: ryankwon@ieee.org; lsw2920@gmail.com)}

\tfootnote{This work was supported by the National Supercomputing Center with supercomputing resources.}

\markboth
{Kwon \headeretal: Enhancing Ship Classification in Optical Satellite Imagery}
{Kwon \headeretal: Enhancing Ship Classification in Optical Satellite Imagery}

\begin{abstract}
In this study, we present an advanced convolutional neural network (CNN) architecture for ship classification based on optical satellite imagery, which significantly enhances performance through the integration of a convolutional block attention module (CBAM) and additional architectural innovations. Building upon the foundational ResNet50 model, we first incorporated a standard CBAM to direct the model's focus toward more informative features, achieving an accuracy of 87\% compared to 85\% of the baseline ResNet50. Further augmentations involved multiscale feature integration, depthwise separable convolutions, and dilated convolutions, culminating in an enhanced ResNet model with improved CBAM. This model demonstrated a remarkable accuracy of 95\%, with precision, recall, and F1 scores all witnessing substantial improvements across various ship classes. In particular, the bulk carrier and oil tanker classes exhibited nearly perfect precision and recall rates, underscoring the enhanced capability of the model to accurately identify and classify ships. Attention heatmap analyses further validated the efficacy of the improved model, revealing more focused attention on relevant ship features regardless of background complexities. These findings underscore the potential of integrating attention mechanisms and architectural innovations into CNNs for high-resolution satellite imagery classification. This study navigates through the class imbalance and computational costs and proposes future directions for scalability and adaptability in new or rare ship-type recognition. This study lays the groundwork for applying advanced deep learning techniques in remote sensing, offering insights into scalable and efficient satellite image classification.
\end{abstract}


\titlepgskip=-21pt

\maketitle

\section{Introduction}
\PARstart{T}{he} advent of high-resolution optical satellite imagery has revolutionized the way we observe and understand the Earth's surface, offering unprecedented opportunities for monitoring, analysis, and decision-making across a wide range of applications. Among these applications, ship classification plays a crucial role in maritime surveillance, traffic management, and environmental monitoring, and provides essential information for ensuring maritime safety, optimizing shipping routes, and preventing illegal activities at sea.

With the proliferation of optical remote sensing (ORS) technologies, the ability to accurately classify ships using satellite images has become increasingly important. However, this task is challenging because of the complex and dynamic nature of the maritime environment, including variations in ship size, type, and appearance, as well as the influence of the sea state, weather conditions, and image resolution.

Recently, convolutional neural networks (CNNs) have emerged as powerful tools for image analysis, demonstrating remarkable success in various fields, including image recognition, object detection, and classification. The deep-learning architecture of CNNs enables them to learn hierarchical feature representations from data, making them particularly well-suited for extracting and classifying intricate patterns in images. This capability has led to the exploration of CNN models for ship classification in optical satellite imagery, aiming to leverage their potential for automatic and accurate identification of ship types from space.

ResNet represent a significant advancement in CNN architecture, introducing the concept of residual learning to address the challenges of training ultra-deep neural networks. By integrating shortcut connections that allow layers to learn residual functions with reference to the input, ResNet facilitate the training of deeper networks, thereby enhancing their capacity for feature representation and improving classification performance. The introduction of ResNet has marked a milestone in deep learning by setting new standards for image recognition and classification tasks\cite{he2016deep}.

Building on the foundation of CNNs and ResNet, the convolutional block attention module (CBAM) introduces an attention mechanism that further refines the feature maps produced by CNNs. By sequentially applying attention across both the spatial and channel dimensions, CBAM enables the network to focus on the most informative features, thereby improving the efficiency and effectiveness of feature representation. This attention-based approach can enhance the performance of CNN architectures across various classification and detection tasks, demonstrating the potential for integrating attention mechanisms into deep-learning models for image analysis\cite{woo2018cbam}.

In this study, we propose a novel framework for ship classification using transfer learning on optical satellite imagery by integrating ResNet with the CBAM to leverage the strengths of both architectures. By combining the depth and residual learning capabilities of ResNet with the selective attention mechanism of the CBAM, our approach aims to achieve superior classification performance by addressing the complexities of ship identification in high-resolution satellite images. Furthermore, we explored the integration of additional enhancements, including multiscale feature integration, depthwise separable convolutions, and dilated convolutions, to further improve the model's capability to accurately capture and classify diverse ship types. Our research contributes to ongoing efforts to harness the power of deep learning for advanced image analysis, offering insights and methodologies that can be applied to various applications beyond maritime surveillance.

\section{Related Work}
Classification of ships using optical satellite imagery is an area of growing interest in the remote sensing community, particularly with the advent of high-resolution imagery and advanced CNNs. Several studies explored different CNN architectures and methodologies to enhance the accuracy and efficiency of ship classification using satellite imagery. This section reviews related work in the domain of ship classification using CNN models, highlighting the significant contributions and methodologies that have shaped current research directions.

Morgan\cite{morgan2019ship} employed the fine-tuning of state-of-the-art CNN architectures for ship detection in remote sensing optical imagery, achieving a notable accuracy of 94.8\% with a fine-tuned Xception model. This study underscores the potential of leveraging pre-existing CNN architectures adapted through fine-tuning to satisfy the specific demands of ship classification tasks. Similarly, Ma et al.\cite{ma2018ship} developed a novel CNN model tailored for marine target classification using GF-3 SAR images, demonstrating its effectiveness at the patch level, and providing a comprehensive scheme for marine target detection in large-scale SAR images. This study demonstrated the adaptability of CNN models to different types of satellite imagery, including synthetic aperture radar (SAR), for ship classification.

Using a different approach, Gadamsetty et al.\cite{gadamsetty2022hash} introduced a deep-learning methodology incorporating hashing with SHA-256 to enhance model integrity and security, facilitating the secure transmission of confidential images with ships detected in satellite imagery. This innovative method highlights the importance of classification accuracy and data security in remote sensing applications. Venkataramani et al.\cite{venkataramani2023harmonizing} trained a CNN-based image segmentation architecture on harmonized optical and SAR satellite images to effectively reduce false positives and negatives in water/non-water classification. Although not directly focused on ship classification, this study demonstrated the utility of CNNs for distinguishing complex features in mixed data sources.

Tienin et al.\cite{tienin2023heterogeneous} introduced spatial-channel attention with a bilinear pooling network (SCABPNet), leveraging a unique dataset combining SAR and optical satellite imagery to enhance feature representation and achieve superior classification performance. The integration of attention mechanisms within CNN architectures, as demonstrated in SCABPNet, illustrates a potential direction for enhancing model sensitivity to relevant features in ship classification tasks. DenseNet-161 was used by Sola et al.\cite{sola2020identifying} to identify sea ice ridging in SAR imagery, achieving a ROC-AUC score of 92.3\% and outperforming previous methods. The application of DenseNet to SAR imagery to detect specific physical phenomena further underscores the versatility of CNN models for remote sensing.

Song et al.\cite{song2021hierarchical} proposed a hierarchical object detection method using an improved saliency detection model to probe suspected regions in complex remote sensing images, followed by using an efficient neural network for precise object categorization and localization. This hierarchical approach, aimed at improving the recall and precision rates for ship detection, demonstrates the potential of combining saliency detection with CNN models to address the challenges of object detection in large-scale optical satellite imagery.

These studies collectively highlighted diverse methodologies and advancements in CNN-based ship classification and detection using optical and SAR satellite imagery. They reflect ongoing efforts to refine CNN architectures, integrate novel data sources, and address the complexities of remote-sensing applications, thereby setting the stage for future research in this rapidly evolving field.

\section{Methodology}
\subsection{Dataset Preparation and Preprocessing}
The cornerstone of our study on ship classification using CNNs is an ORS ship dataset. This dataset comprises eight distinct ship classes: bulk carriers, car carriers, cargo, chemical tankers, containers, dredges, oil tankers, and tugs, encompassing 8678 images. These images were meticulously collected via Google Earth to ensure submeter resolution and to capture the intricate details of the ships.

Given the diverse and comprehensive nature of the ORS dataset, it is an ideal testbed for evaluating the efficacy of deep-learning models in classifying high-resolution satellite imagery into meaningful ship categories. However, certain preparatory and preprocessing steps are necessary to optimize the dataset for our experimental framework.

Preprocessing Steps:

\begin{enumerate}
    \item Data Cleaning: The initial step involves removing potential anomalies or irrelevant entries from the dataset. This included verifying the integrity of the image data and ensuring that each class label accurately reflects the image content.

    \item Dataset Segmentation: Each image within the dataset was carefully segmented to ensure that it contained only one ship. This segmentation process is critical for minimizing background noise and focusing the model's learning on ships, thereby reducing the complexity of the classification task.

    \item Class Selection: Given the uneven distribution of images across the eight ship classes, classes were excluded where the test set, based on an 80:20 train-test split, contained 100 or fewer images. This criterion was applied to maintain balance in the dataset and ensure that the models were trained and tested on sufficiently representative samples from each class.

    \item Train-Test Split: The dataset was split into training and testing sets in an 80:20 ratio, ensuring that the models had access to a diverse and comprehensive set of images for learning while reserving a substantial subset for evaluation purposes.

    \item Image Resizing and Normalization: To accommodate the input requirements of the CNN models, all images were resized to a uniform dimension of 224x224 pixels. This resizing step was accompanied by normalization, adjusting pixel values to a common scale to facilitate more stable and faster convergence during model training.

    \item Data Augmentation: To enhance the robustness and generalizability of the model across various imaging conditions, data augmentation techniques such as random horizontal flips, random vertical flips, and random rotations were applied. This step not only expands the diversity of the training data but also simulates a broader range of imaging scenarios that the model might encounter in real-world applications.
\end{enumerate}

Through these preprocessing steps, we aimed to create a well-structured, balanced, and comprehensive dataset that would enable effective training and evaluation of our CNN models for ship classification. The meticulous preparation of the ORS dataset underscores our commitment to leveraging high-quality real-world data to advance state-of-the-art maritime object detection and classification using optical satellite imagery.

\subsection{Model Architecture and Training}
Our research aims to push the boundaries of ship classification from optical satellite imagery by deploying advanced CNN models. The core of our methodology is to leverage a modified ResNet50 architecture, augmented with a CBAM, and further enhanced with additional architectural innovations, such as multiscale feature integration, depthwise separable convolutions, and dilated convolutions. These modifications were designed to increase the accuracy and efficiency of the model, enabling it to capture various features from complex satellite images.

\subsubsection{ResNet50-based Transfer Learning}
We commenced with the ResNet50 model, which is renowned for its residual learning framework and crucial for training deep networks. By incorporating shortcut connections, ResNet50 facilitates the training process, allowing the network to learn residual functions with reference to the input, rather than unreferenced functions. This foundational model was selected for its proven track record in image recognition tasks, providing a solid basis for our modifications.

\begin{figure}[htbp]
    \centering
    \includegraphics[width=0.4\linewidth]{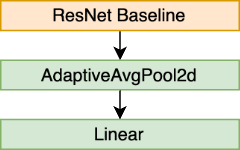}
    \caption{ResNet50-based Model Architecture}
    \vspace{-1em}
\end{figure}

\subsubsection{Convolutional Block Attention Module (CBAM)}
To further refine the feature maps produced by ResNet50, we integrate a CBAM into our architecture. CBAM systematically infers attention maps along both the channel and spatial dimensions, enabling the network to focus adaptively on more informative features. This attention mechanism enhances the interpretability and efficiency of the model for processing satellite imagery, leading to an improved classification performance.

\begin{figure}[htbp]
    \centering
    \includegraphics[width=0.7\linewidth]{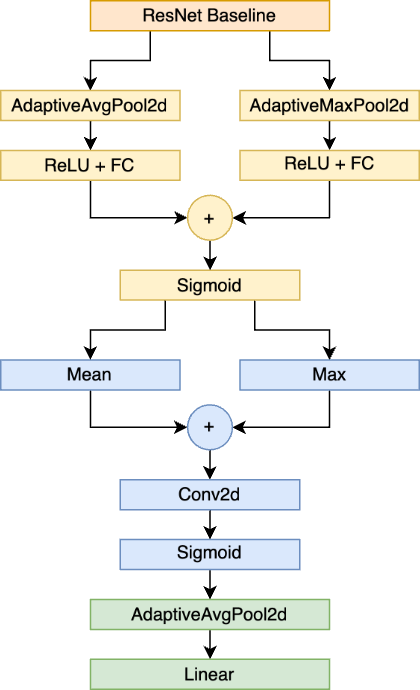}
    \caption{ResNet-CBAM Integrated Model Architecture}
    \vspace{-1em}
\end{figure}

\subsubsection{Training and Evaluation}
The training process was conducted using a dataset split into training and testing subsets, with the images resized to 224 × 224 pixels and normalized. We employed data augmentation techniques such as random horizontal flips and rotations to increase the diversity of the training data. The model was trained using cross-entropy loss, optimized with the Adam optimizer, with the learning rate set to 1e-4, and adjusted dynamically based on performance metrics.

The effectiveness of the proposed modifications was evaluated through rigorous testing, including quantitative assessments of the classification accuracy, precision, recall, and F1 scores. A comparative analysis against baseline models and state-of-the-art approaches in ship classification demonstrates the superiority of our model in handling satellite imagery complexities.

\subsection{Evaluation and Performance Analysis}
To advance ship classification using optical satellite imagery, the efficacy of our proposed CNN models was rigorously assessed using a comprehensive evaluation and performance analysis framework. Our evaluation methodology was based on a comparative analysis, juxtaposing the baseline ResNet50 model, ResNet50 integrated with a CBAM in its standard form, and our enhanced ResNet model featuring an improved CBAM with additional architectural innovations. This comparative approach allowed us to meticulously assess the incremental benefits introduced by each modification, thereby validating the effectiveness of the proposed enhancements.

\subsubsection{Evaluation Metrics}
The performance of each model variant was evaluated against a set of widely recognized metrics in the image classification domain.

\begin{itemize}
    \item Accuracy: The proportion of correctly classified instances out of the total number of instances provides a high-level view of the overall performance of the model.
    \item Precision and Recall: These metrics offer insights into the model's ability to correctly identify positive instances while minimizing false positives, which are essential for applications requiring high confidence in classification results.
    \item F1-Score: The harmonic mean of precision and recall, offering a balanced measure that considers both precision and recall of the model. This is particularly useful when dealing with unbalanced datasets.
    \item Confusion Matrix: A detailed breakdown of the model's performance across different classes provides insights into specific areas of strengths and weaknesses in classification.
\end{itemize}

\subsubsection{Comparative Analysis Methodology}
The comparative analysis was structured as follows:

\begin{enumerate}
    \item Baseline Model (ResNet50): Initially, the performance of the unmodified ResNet50 model served as the baseline. This allowed us to establish a reference point for assessing the incremental improvements brought about by integrating the CBAM and our subsequent enhancements.
    \item ResNet50 + Standard CBAM: The second phase involved evaluating the ResNet50 model integrated with standard CBAM. This step aimed to quantify the impact of adding an attention mechanism to the baseline model, focusing on how attention to spatial and channel features could improve classification accuracy.
    \item Enhanced ResNet Model with Improved CBAM: Finally, we assessed our enhanced ResNet model, which not only incorporated CBAM but also introduced additional architectural innovations aimed at optimizing the model for satellite imagery classification. This includes multiscale feature integration, depth-wise separable convolutions, and dilated convolutions.
\end{enumerate}

\subsubsection{Performance Analysis Process}
For each model variant, we conducted a series of experiments under identical conditions to ensure fair comparison. The ORS dataset, which was prepared and preprocessed as described in the previous sections, served as the basis for training and testing across all models. Data augmentation techniques have been consistently applied to mitigate overfitting and enhance the generalization capabilities of models.

Following training, each model was evaluated using the designated testing dataset, and the results were captured across the specified metrics. This process facilitates a nuanced understanding of the capabilities and limitations of each model, particularly for ship classification using high-resolution optical satellite imagery.

\section{Experiments and Results}
\subsection{Experimental Setup}
A rigorous experimental framework was established to assess the performance of the proposed models for advancing ship classification using CNNs. This section describes the specifics of our experimental setup, ensuring reproducibility and providing clarity on the conditions under which the experiments were conducted.

\subsubsection{Dataset and Preprocessing}
The experiments utilized an ORS ship dataset, meticulously prepared as previously described. This dataset encompasses various ship types, each represented by high-resolution satellite imagery. After preprocessing, the final dataset included images across four main ship classes, and data augmentation techniques were applied to enhance model robustness and generalizability. The dataset was split into training and testing sets in an 80:20 ratio to ensure a comprehensive evaluation of the model performance.

\subsubsection{Hardware and Software Configuration}
The experiments were executed on a high-performance computing cluster equipped with the following specifications to accommodate the computational demands of training the deep learning models:

\begin{itemize}
    \item CPU: An AMD EPYC 7R32 processor was used to efficiently manage dataset loading, preprocessing tasks, and model training orchestration.
    \item GPU: Deployment of an NVIDIA A10G graphics processing unit, offering substantial computational power and 24GB of memory to facilitate rapid model training and experimentation.
    \item Memory: Accesses 128 GB of system memory, ensuring sufficient capacity for handling large datasets, model parameters, and intermediate computations.
    \item Operating System: Ubuntu 20.04.1, a stable and widely supported Linux distribution that ensures a reliable and consistent software environment.
\end{itemize}

\subsubsection{Deep Learning Framework}
Our models were developed and trained using PyTorch, a flexible and powerful deep-learning library that favors dynamic computation graphs and extensive model development ecosystems. PyTorch's support for CUDA enabled the leveraging of GPU acceleration, significantly enhancing training efficiency and experimental throughput.

\subsubsection{Evaluation Metrics}
The model performance was quantitatively evaluated using a suite of metrics, including accuracy, precision, recall, F1-score, and the construction of confusion matrices. These metrics were chosen because of their ability to provide a comprehensive overview of the model efficacy, from the overall accuracy to the balance between precision and recall.

\subsection{Training Details}
To ensure a fair and thorough evaluation of each model's performance in the ship classification task using the ORS dataset, we designed a training regimen. This section outlines the training details, including the specific configurations and methodologies employed for each of the model variants under investigation: the baseline ResNet50, ResNet50 integrated with a standard CBAM, and our enhanced ResNet model with improved CBAM and additional architectural innovations.

\subsubsection{Training Configuration}
All models were trained using the following unified configuration to facilitate an equitable comparison:

\begin{itemize}
    \item Batch Size: Set to 128, this size was chosen to balance the trade-off between memory utilization and the stability of gradient updates.
    \item Epochs: Each model was trained for 30 epochs. This duration was determined to be sufficient for the models to converge based on preliminary experiments.
    \item Learning Rate: Initiated at 1e-4 for all models. This learning rate was selected to ensure steady convergence without overshooting the minima in the landscape loss.
    \item Optimizer: The Adam optimizer was used for its adaptive learning rate capabilities, aiding in the efficient convergence of the models.
    \item Learning Rate Scheduler: A step-decay learning rate scheduler was employed, reducing the learning rate by a factor of 0.1 every 10 epochs to fine-tune the models as they approached convergence.
    \item Loss Function: Cross-entropy loss was used as the primary criterion for training because of its effectiveness in classification tasks and its ability to handle multiple classes smoothly.
\end{itemize}

\subsubsection{Data Augmentation}
To enhance the ability of the model to generalize unseen data and mitigate the risk of overfitting, data augmentation techniques were applied to the training dataset. These included random horizontal flips, rotations within 10 °, and vertical flips. These transformations introduce variability into the training process, simulating various imaging conditions without requiring additional labeled data.

\subsubsection{Training Environment}
Training was conducted using PyTorch on a high-performance computing setup, as detailed in the "Experimental Setup" section. Using an NVIDIA A10G GPU allowed for efficient training, significantly reducing the time required to train each model variant.

\subsubsection{Model Evaluation}
Model performance was continuously monitored throughout the training process using a validation set derived from the original training data. This validation set, comprising 20\% of the training data, enabled the early identification of overfitting and assessment of model generalizability. Model checkpoints were saved at the end of each epoch and the final model selection was based on the best performance of the validation set across all epochs.

\subsection{Performance of ResNet50 with Standard CBAM}
Integrating the CBAM into the ResNet50 architecture is a significant step toward enhancing the model's ability to focus on relevant features for ship classification. The addition of the CBAM improved the model's precision, recall, and F1 score across all ship classes compared with the baseline ResNet50 model. Specifically, the precision and recall for the bulk carrier class show noticeable improvements, indicating an enhanced capability to distinguish this class from the others. The overall accuracy of the model increased to 0.87, from 0.85 in the baseline model, underscoring the effectiveness of incorporating attention mechanisms into CNNs for satellite imagery classification (Figure 1). The weighted average precision and recall across all classes also improved, confirming the balanced performance enhancement of the model.

\begin{table}[htbp!]
\centering
\begin{tabularx}{\linewidth}{l *{4}{X}}
\toprule
\textbf{Class} & \textbf{Precision} & \textbf{Recall} & \textbf{F1-Score} & \textbf{Support} \\
\midrule
\textbf{Bulk Carrier} & 0.83 & 0.81 & 0.82 & 411 \\
\textbf{Cargo} & 0.93 & 0.90 & 0.91 & 308 \\
\textbf{Container} & 0.85 & 0.76 & 0.80 & 258 \\
\textbf{Oil Tanker} & 0.88 & 0.94 & 0.91 & 691 \\
\midrule
\textbf{Accuracy} &  &  & 0.87 & 1668 \\
\textbf{Macro Avg.} & 0.87 & 0.85 & 0.86 & 1668 \\
\textbf{Weighted Avg.} & 0.87 & 0.87 & 0.87 & 1668 \\
\bottomrule
\end{tabularx}
\caption{Performance Metrics of ResNet-Standard CBAM}
\label{table:performance_metrics_1}
\vspace{-2em}
\end{table}

\subsection{Enhanced ResNet Model with Improved CBAM Performance}
The enhanced ResNet model with improved CBAM demonstrated a notable leap in performance metrics, highlighting the substantial impact of additional architectural innovations. The model achieved a remarkable overall accuracy of 95\%, with significant improvements in precision and recall across all ship classes. In particular, the bulk carrier and oil tanker classes exhibited nearly perfect precision and recall, indicating the exceptional ability of the model to accurately identify and classify these ships. The macro and weighted averages for precision, recall, and F1-score exceeded those of both the baseline ResNet50 and ResNet50 with standard CBAM models, highlighting the value of our enhancements in handling the intricacies of satellite-based ship classification (Figure 2).

A comparative analysis of these models revealed the incremental benefits conferred by each modification layer. The enhanced ResNet model with improved CBAM not only capitalized on the strengths of CBAM but also leveraged multiscale feature integration, depthwise separable convolutions, and dilated convolutions to address the specific challenges posed by high-resolution satellite imagery. This comprehensive approach resulted in a model that not only excels in classification accuracy but also demonstrates a nuanced understanding of the spatial and feature-specific complexities inherent in satellite images of maritime vessels.

\begin{table}[htbp!]
\centering
\begin{tabularx}{\linewidth}{l *{4}{X}}
\toprule
\textbf{Class} & \textbf{Precision} & \textbf{Recall} & \textbf{F1-Score} & \textbf{Support} \\
\midrule
\textbf{Bulk Carrier} & 0.94 & 0.95 & 0.94 & 405 \\
\textbf{Cargo} & 0.94 & 0.93 & 0.94 & 330 \\
\textbf{Container} & 0.91 & 0.90 & 0.90 & 254 \\
\textbf{Oil Tanker} & 0.98 & 0.98 & 0.98 & 679 \\
\midrule
\textbf{Accuracy} &  &  & 0.95 & 1668 \\
\textbf{Macro Avg.} & 0.94 & 0.94 & 0.94 & 1668 \\
\textbf{Weighted Avg.} & 0.95 & 0.95 & 0.95 & 1668 \\
\bottomrule
\end{tabularx}
\caption{Performance Metrics of Improved CBAM}
\label{table:performance_metrics_2}
\vspace{-2em}
\end{table}

\subsubsection{Attention Heatmaps Analysis}
\begin{figure*}
    \centering
    \includegraphics[width=1\textwidth]{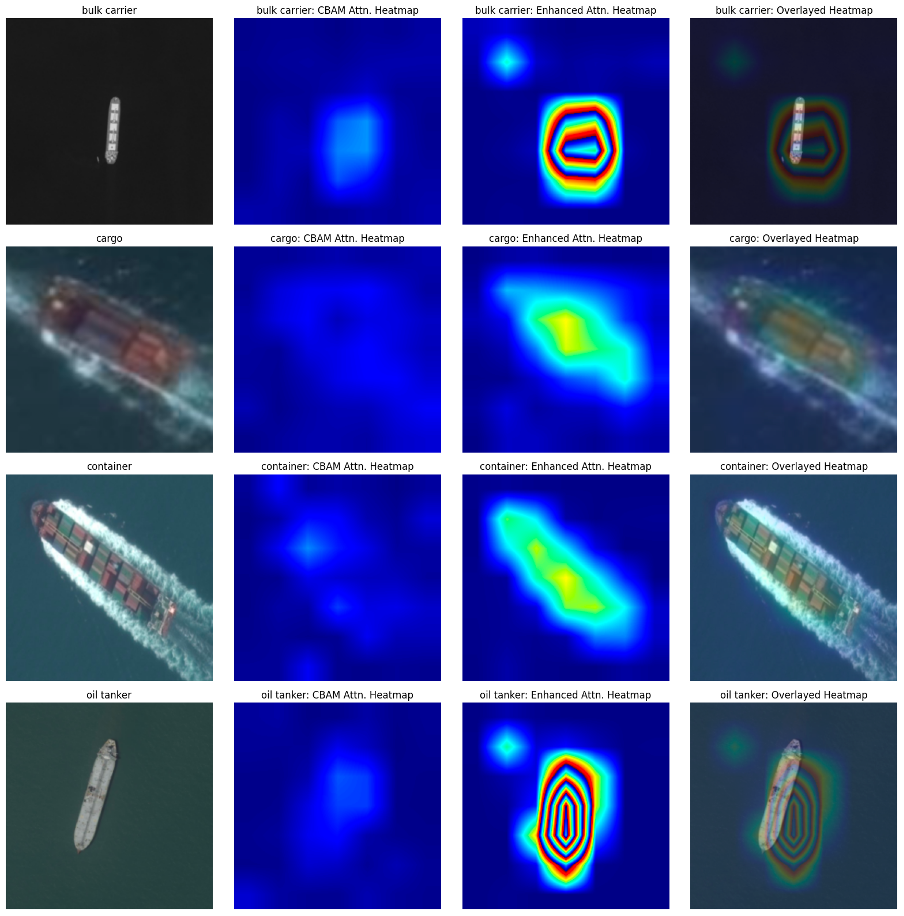}
    \caption{Heatmaps highlight the enhanced CBAM's ability to focus on key features for accurate ship classification}
    \label{fig:heatmaps}
\end{figure*}

To further understand the impact of the enhanced CBAM on the model performance, attention heatmaps were generated for a subset of the test images. These heat maps provide visual evidence of the focus areas of the model, revealing that the improved CBAM enables the model to concentrate more effectively on ships within the images, irrespective of their size, orientation, or background complexity. The heat maps underscore the capability of the model to discern subtle features indicative of different ship classes, a testament to the efficacy of the enhanced attention mechanism (Figure \ref{fig:heatmaps}).

\subsection{Figures and Confusion Matrices}
\begin{figure}
    \centering
    \includegraphics[width=1\linewidth]{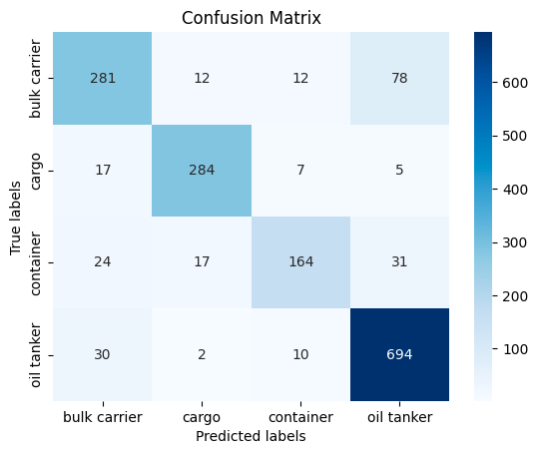}
    \caption{Confusion matrix of ResNet50 baseline model}
    \label{fig:conf_mtrx_1}
\end{figure}

\begin{figure}
    \centering
    \includegraphics[width=1\linewidth]{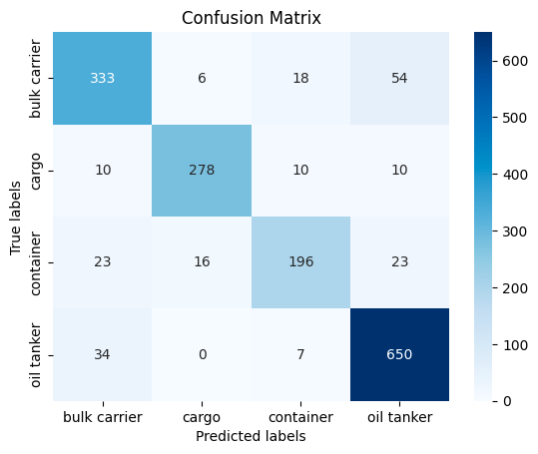}
    \caption{Confusion matrix of ResNet50 + Standard CBAM}
    \label{fig:conf_mtrx_2}
\end{figure}

\begin{figure}
    \centering
    \includegraphics[width=1\linewidth]{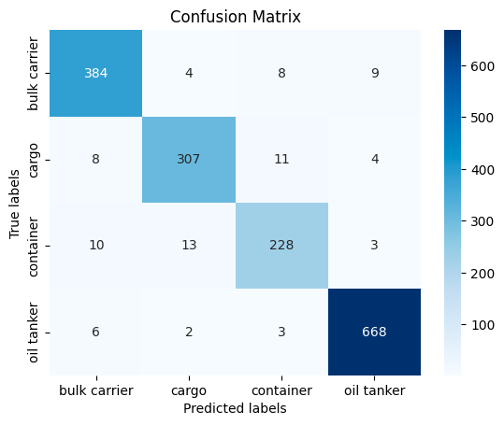}
    \caption{Confusion matrix of enhanced ResNet with improved CBAM}
    \label{fig:conf_mtrx_3}
\end{figure}

The performance differences between the models are visually represented in the graphs, and confusion matrices are provided. These figures illustrate not only the quantitative advancements achieved by each subsequent model iteration but also the qualitative improvements in the model's classification logic and attention to detail. In particular, the confusion matrices highlight the reduction in misclassifications and increasing reliability of the models in correctly identifying ship types, with the enhanced ResNet model with improved CBAM exhibiting the most significant advancements (Figures \ref{fig:conf_mtrx_1}, \ref{fig:conf_mtrx_2}, and \ref{fig:conf_mtrx_3}).

\section{Discussion of Findings and Limitations}
Experimental results presented a compelling narrative regarding the effectiveness of integrating attention mechanisms and architectural innovations in CNN models for classifying ships using optical satellite imagery. The enhanced ResNet model with improved CBAM notably outperformed the baseline ResNet50 and ResNet50 models integrated with standard CBAM in terms of accuracy, precision, recall, and F1 score across all ship classes. This improvement underscores the value of attention mechanisms, such as CBAM, which enables the model to focus on the most informative features of an image, thus improving the classification performance. Furthermore, the incorporation of multiscale feature integration, depthwise separable convolutions, and dilated convolutions addresses the specific challenges posed by high-resolution satellite images, such as scale variations and complex backgrounds.

The attention heat map analysis provided additional insights into the performance of the model, visually demonstrating how the improved CBAM guided the model to focus on relevant features for classification. This capability is critical for the accurate classification of ships in satellite imagery, where the presence of diverse features and noise can significantly affect the performance.

\subsection{Limitations}
The experimental results presented a compelling narrative regarding the effectiveness of integrating attention mechanisms and architectural innovations in CNN models for the classification of ships using optical satellite imagery. The enhanced ResNet model with improved CBAM notably outperformed the baseline ResNet50 and ResNet50 models integrated with standard CBAM in terms of accuracy, precision, recall, and f1-score across all ship classes. This improvement underscores the value of attention mechanisms, such as CBAM, which enables the model to focus on the most informative features of an image, thus improving the classification performance. Furthermore, the incorporation of multiscale feature integration, depthwise separable convolutions, and dilated convolutions addresses the specific challenges posed by high-resolution satellite images, such as scale variations and complex backgrounds.

The attention heat map analysis provided additional insights into the model performance, visually demonstrating how the improved CBAM guided the model to focus on relevant features for classification. This capability is critical for the accurate classification of ships in satellite imagery, where the presence of diverse features and noise can significantly affect the performance.

\subsection{Challenges}
A recurring challenge in satellite image classification is the variability in image quality and conditions such as lighting, weather, and seasonal changes. Although the proposed model demonstrated robustness against these variations, further work is required to enhance its adaptability to extreme conditions.

Moreover, the reliance on labeled datasets for training poses a bottleneck to scalability and adaptability to new or rare ship types. The labor-intensive process of labeling satellite imagery limits dataset size and diversity, potentially hindering the comprehensive understanding of global maritime traffic by the model.

\section{Conclusion}
The findings of this study highlight the potential of advanced CNN architectures augmented with attention mechanisms and architectural innovations for the classification of ships in optical satellite imagery. The enhanced ResNet model with improved CBAM represents a significant step forward in this domain, offering improved accuracy and a deeper understanding of the visual features crucial for classification.

Addressing the identified limitations and challenges is crucial for further advancement. This includes exploring strategies to mitigate class imbalances, reduce computational costs, and enhance the model's robustness against variable image conditions. Additionally, leveraging unsupervised or semi-supervised learning techniques can alleviate the dependency on large labeled datasets, paving the way for more scalable and adaptable satellite image classification models.

\bibliographystyle{unsrt}
\bibliography{_references}

\EOD

\end{document}